\documentclass[letterpaper, 10 pt, conference]{ieeeconf}  % Comment this line out
\IEEEoverridecommandlockouts                              % This command is only
\usepackage{graphicx} % for pdf, bitmapped graphics files
\usepackage{amsmath} % assumes amsmath package installed
\usepackage{amssymb}  % assumes amsmath package installed
\usepackage{cite} % Recommended for IEEE style citations
\usepackage[dvipsnames]{xcolor}

\usepackage{tikz}
\usepackage{adjustbox}
\usetikzlibrary{
  arrows.meta,
  positioning,
  shapes.geometric,
  calc
}
\tikzset{
  box/.style={draw, rounded corners, minimum width=2cm, minimum height=1cm, align=center},
  smallbox/.style={draw, rounded corners, minimum width=1.6cm, minimum height=0.8cm, align=center},
  circleop/.style={draw, circle, minimum size=6mm},
  arrow/.style={->, thick},
  net/.style={draw, rectangle, minimum width=2.8cm, minimum height=2.5cm},
}
\newcommand{\Param}{\mathbf{f}}

\tikzset{BasisFNN_net/.pic={
    \node[draw, circle, fill] (il1) at (-1, 1) {};
    \node[draw, circle, fill] (il2) at (-1, 0) {};
    \node[draw, circle, fill] (il3) at (-1, -1) {};

    \node[draw, circle, fill] (hl1) at (0, 1) {};
    \node[draw, circle, fill] (hl2) at (0, 0) {};
    \node[draw, circle, fill] (hl3) at (0, -1) {};
    
    \foreach \start in {il1, il2, il3}
    {
        \foreach \stop in {hl1, hl2, hl3}
            \draw[-stealth] (\start) -- (\stop);
    }        
    
    \node[draw, circle, fill] (ol1) at (1, 1) {};
    \node[draw, circle, fill] (ol2) at (1, 0) {};
    \node[draw, circle, fill] (ol3) at (1, -1) {};

    \foreach \start in  {hl1, hl2, hl3}
    {
        \foreach \stop in {ol1, ol2, ol3}
            \draw[-stealth] (\start) -- (\stop);
    }
}
}
\tikzset{enc_net/.pic={
    \node[draw, circle, fill] (il1) at (-1, 0.5) {};
    \node[draw, circle, fill] (il2) at (-1, -0.5) {};

    \node[draw, circle, fill] (hl1) at (0, 1) {};
    \node[draw, circle, fill] (hl2) at (0, 0) {};
    \node[draw, circle, fill] (hl3) at (0, -1) {};
    
    \foreach \start in {il1, il2}
    {
        \foreach \stop in {hl1, hl2, hl3}
            \draw[-stealth] (\start) -- (\stop);
    }        
    
    \node[draw, circle, fill] (ol1) at (1, 1) {};
    \node[draw, circle, fill] (ol2) at (1, 0) {};
    \node[draw, circle, fill] (ol3) at (1, -1) {};

    \foreach \start in  {hl1, hl2, hl3}
    {
        \foreach \stop in {ol1, ol2, ol3}
            \draw[-stealth] (\start) -- (\stop);
    }
}
}
\tikzset{dec_net/.pic={
    \node[draw, circle, fill] (il1) at (-1, 1) {};
    \node[draw, circle, fill] (il2) at (-1, 0) {};
    \node[draw, circle, fill] (il3) at (-1, -1) {};
    
    \node[draw, circle, fill] (hl1) at (0, 1) {};
    \node[draw, circle, fill] (hl2) at (0, 0) {};
    \node[draw, circle, fill] (hl3) at (0, -1) {};
    
    \foreach \start in {il1, il2, il3}
    {
        \foreach \stop in {hl1, hl2, hl3}
            \draw[-stealth] (\start) -- (\stop);
    }        
    
    \node[draw, circle, fill] (ol1) at (1, 0.5) {};
    \node[draw, circle, fill] (ol2) at (1, -0.5) {};

    \foreach \start in  {hl1, hl2, hl3}
    {
        \foreach \stop in {ol1, ol2}
            \draw[-stealth] (\start) -- (\stop);
    }
}
}

\tikzset{rnn_node/.pic={
    \node[draw, circle, fill] (il1) at (-1, 1) {};
    \node[draw, circle, fill] (il2) at (-1, 0) {};
    \node[draw, circle, fill] (il3) at (-1, -1) {};
    
    \node[draw, circle, fill] (hl1) at (0, 1) {};
    \node[draw, circle, fill] (hl2) at (0, 0) {};
    \node[draw, circle, fill] (hl3) at (0, -1) {};
    
    \foreach \start in {il1, il2, il3}
    {
        \foreach \stop in {hl1, hl2, hl3}
            \draw[-stealth] (\start) -- (\stop);
    }        
    
    \node[draw, circle, fill] (ol1) at (1, 1) {};
    \node[draw, circle, fill] (ol2) at (1, 0) {};
    \node[draw, circle, fill] (ol3) at (1, -1) {};

    \foreach \start in  {hl1, hl2, hl3}
    {
        \foreach \stop in {ol1, ol2, ol3}
            \draw[-stealth] (\start) -- (\stop);
    }
}
}

\tikzset{>={Latex[width=2mm, length=3mm]}}
\newcommand{\Reals}{\mathbb{R}}
\newcommand{\OpF}{\mathtt{N}}
\newcommand{\OpG}{\mathcal{G}}

\newcommand{\bphi}{\boldsymbol{\phi}}

\definecolor{blue}{HTML}{0072BD}
\definecolor{orange}{HTML}{D95319}
\definecolor{green}{HTML}{77AC30}
\definecolor{purple}{HTML}{7E2F8E}

\usepackage[caption=false,font=footnotesize]{subfig}
\usepackage{multirow} % put in preamble
\usepackage{hyperref}
\usepackage{makecell}
\usepackage{booktabs}

\title{
    RHYME-XT: A Neural Operator for Spatiotemporal Control Systems
}

\author{Marijn Ruiter$^{1}$, Miguel Aguiar$^{1}$, Jake Rap$^{2}$, Karl H. Johansson$^{1}$,  Amritam Das$^{2}$ % <-this % stops a space
\thanks{$^{1}$ Digital Futures and Division of Decision and Control Systems, KTH Royal Institute of Technology, SE-100 44 Stockholm, Sweden. Email: {\tt\small marijnruiter09@gmail.com, $\{$aguiar,kallej$\}$@kth.se}
$^{2}$ Control Systems Group, EE Dept., Eindhoven University of Technology, P.O. Box 513, 5600 MB Eindhoven, The Netherlands, Email: {\tt\small $\{$j.e.w.rap, am.das$\}$@tue.nl}}}%

\begin{document}

\maketitle
\thispagestyle{empty}
\pagestyle{empty}

%%%%%%%%%%%%%%%%%%%%%%%%%%%%%%%%%%%%%%%%%%%%%%%%%%%%%%%%%%%%%%%%%%%%%%%%%%%%%%%%
\begin{abstract}
We propose RHYME‑XT, an operator-learning framework for surrogate modeling of spatiotemporal control systems governed by input‑affine nonlinear partial integro‑differential equations (PIDEs) with localized rhythmic behavior. RHYME‑XT uses a Galerkin projection to approximate the infinite-dimensional PIDE on a learned finite-dimensional subspace with spatial basis functions parameterized by a neural network. This yields a projected system of ODEs driven by projected inputs. Instead of integrating this non-autonomous system, we directly learn its flow map using an architecture for learning flow functions, avoiding costly computations while obtaining a continuous-time and discretization‑invariant representation.
Experiments on a neural field PIDE show that RHYME‑XT outperforms a state‑of‑the‑art neural operator and is able to transfer knowledge effectively across models trained on different datasets, through a fine-tuning process.
\end{abstract}

%%%%%%%%%%%%%%%%%%%%%%%%%%%%%%%%%%%%%%%%%%%%%%%%%%%%%%%%%%%%%%%%%%%%%%%%%%%%%%%%
\section{Introduction}

Dynamical systems that are distributed over space and time are typically modeled using partial differential equations (PDEs) or partial integro-differential equations (PIDEs). When the PDEs or PIDEs are nonlinear, the dynamical system may have more than one equilibrium with unique properties (e.g., stable, unstable, limit cycle) depending on the initial condition. For example, the behavior can be in the form of shock waves in fluids \cite{garriga_simulation_2021} or self-sustained oscillations in transmission lines \cite{wu_self-sustained_1996}.
When such a model is driven by external inputs alongside initial conditions, depending on the input energy, the behavior may switch from one behavioral regime to another.
In the literature, such a phenomenon is referred to as \emph{excitability} \cite{SepulchreEtAl18}.
Excitable behavior is particularly exemplified by neuronal dynamics, where small inputs result in a small output, but once an activation threshold is met, the output changes rapidly, resulting in spike trains and oscillatory behavior \cite{Sepulchre22, Amari77}. 

Solving nonlinear PIDEs and PDEs is challenging.
Classical approaches consist of simulating them using numerical discretization methods, such as finite element or volume methods. However, due to the stiff and local behavior of these systems, the accuracy depends on the choice of discretization scheme for the spatial and temporal domains.
This means that simulations on fine grids are accurate but computationally expensive, whereas simulations on coarse grids are faster, but inaccurate and possibly unstable.

Model-order reduction can circumvent this curse of dimensionality and reduce the computational complexity. Projection-based methods, such as the Galerkin method, have seen successful application in the numerical simulation of fluid dynamics \cite{RowleyEtAl04}. 
However, these methods still rely on solving an albeit lower-dimensional system of differential equations and require explicit knowledge of the governing equations.

An alternative approach is to use data to learn a surrogate (approximate) model of the underlying spatiotemporal behavior.
One such approach is based on the Koopman operator for lifting the dynamics to a higher-dimensional space in which their evolution is linear \cite{BevandaEtAl21}.
However, this approach is mostly restricted to finite-dimensional control-affine systems, though extensions to more general systems have been proposed in~\cite{KordaMezic18}.
Alternatively, physics-informed neural networks (PINNs) \cite{KarniadakisEtAl21} can also learn surrogate models. However, PINNs are trained for a single scenario (e.g., fixed initial condition).
Neural operators, a set of architectures based on the capability of neural networks to approximate infinite-dimensional operators, do not suffer from this limitation, and have been applied in control settings \cite{BhanEtAl23}.
However, they require more extensive training data from varying scenarios.
Flow approximation of control systems has recently been developed for the simulation of spiking systems~\cite{AguiarEtAl24}.
This architecture is based on discrete-time recurrent neural networks (RNN) and is a universal approximator of flows of continuous-time dynamical systems with inputs~\cite{AguiarEtAl23}.

Our main contribution in this paper is RHYME-XT, an operator learning framework for modeling the solution operator mapping a spatiotemporal initial condition and input function to the corresponding trajectory of spatiotemporal dynamics described by a nonlinear, input-dependent PIDE.
We evaluate our method on a dynamical system consisting of a network of neurons, described by the neural field equation, and compare it against a baseline architecture, showing improved ability to handle the nonlinear dynamics associated with the system.
Furthermore, we demonstrate that a model trained on a large dataset can be efficiently fine‑tuned to approximate a different model of the same family, with only a fraction of the data previously required.

The method is based on extending the flow learning method introduced in \cite{AguiarEtAl23} to nonlinear infinite-dimensional systems.
Compared to the existing literature, our proposed architecture is discretization-free (both in space and time) and requires neither prior selection of basis functions nor any knowledge of the underlying dynamics.
Specifically, using a learned basis parameterized by a feedforward neural network (FNN), we project the dynamics onto a finite-dimensional subspace.
The projected model is directly approximated using an RNN. The output is then reconstructed using a nonlinear combination of these learned bases and temporal coefficients. In this way, we leverage the approximation and learning capabilities of neural networks, while simultaneously ensuring a continuous representation of the solution. As this parallel modeling bears semblance to the sub-net structure introduced in the Deep Operator Network (DeepONet) \cite{LuEtAl21}, we consider that as our baseline.
Our method shows superior results, and the resulting model is able to make predictions beyond the temporal horizon used in the training phase. Additionally, we consider a transfer learning strategy, enabling the transfer of knowledge among models trained on distinct datasets.

The paper is organized as follows. In section \ref{sec: problem formulation}, we present the considered family of PIDEs and define the operator we are interested in learning. Section~\ref{sec: finite dimension} details the PIDE approximation using the Galerkin method. In section~\ref{sec: model approximation with Galerkin Projection}, we introduce the novel neural operator architecture. Section~\ref{sec: Learning-based Approach} details how the neural network parameterizations are learned. Section~\ref{sec: Results} showcases the numerical results on a network of neurons. Finally, section \ref{sec: Conclusion} concludes with some perspectives on future research directions.

\section{Problem Formulation} \label{sec: problem formulation}

Consider the following class of input-affine nonlinear PIDEs
\begin{equation} \label{eq:PIDEsfamily}
\begin{aligned}
    \partial_t u(x, t) &= \OpF[u(\cdot, t)](x) + f(x,t), \\
    u(x,0) &= u_0(x),  
\end{aligned}
\end{equation}
where \begin{math}
        x \in \Omega \subset \Reals^d
\end{math} and \begin{math} 
        t \in [0, T]
\end{math} represent space and time, respectively.
%and \begin{math} \Omega \end{math} and \begin{math} [0, T] \end{math} their domains.
The function \begin{math}
        u : \Omega \times [0, T] \to \Reals
\end{math} is the physical quantity of interest, with the initial value (at $t = 0$)
\begin{math}
        {u_0: \Omega \to \Reals}
\end{math}. Furthermore, \begin{math}
       \OpF 
\end{math} is a possibly nonlinear operator and \begin{math}
        f: \Omega \times [0, T] \to \Reals
\end{math} is the input function.

An example of such a system is the single-layer neural field model proposed in \cite{Amari77} described as the integral equation
\begin{equation} \label{eq:neural_field}
    \partial_t u(x,t) = f(x,t)-u(x,t) + \int_{\Omega} k(|x-y|)  {h}{\left(u(y,t) -\theta\right)} \mathrm{d}y,
\end{equation}
where $u$ is the membrane potential, $h$ is a function that dictates the firing rate,  $\theta$ is the activation threshold, $k$ is a weighting kernel representing synaptic connectivity between neurons at locations $x$ and $y$, and $f$ is a stimulus that we regard as an input.

Consider the operator 
\begin{equation}
    \OpG: (u_0, f) \rightarrow u
\end{equation}
mapping the initial condition and input function to the corresponding solution of \eqref{eq:PIDEsfamily} on the domain $\Omega \times [0,T]$.
We make the standing assumption that $\OpG$ is well defined, i.e. the existence and uniqueness of a weak solution $u$ for a given $u_0$ and $f$.
We are interested in learning an approximation $\hat{\OpG}$ of $\OpG$ minimizing
\begin{equation}
    \mathcal{L}({\hat{\OpG}}) =  \frac{1}{T|\Omega|} \mathbf{E} \int_\Omega \int_0^T \bigl\|\OpG(u_0,f) - \hat{\OpG}(u_0, f)\bigr\|_2^2 \,\mathrm{d}x \,\mathrm{d}t,
\end{equation}
where the expectation is taken over $u_0$ and $f$, whose distributions 
are user-defined and problem-dependent.
Furthermore, we wish to learn $\hat{\OpG}$ from data consisting of
$N$ trajectories of~\eqref{eq:PIDEsfamily}:
\begin{equation}\label{eq:data}
\left( (u_0^{(i)}, f^{(i)}), u^{(i)} \right) , ~ i=1, \dots , N,
\end{equation}
with $u^{(i)}=\OpG(u_0^{(i)},f^{(i)})$.
In practice, we have access to data only at a finite set of evaluation points $S_i=\{(x_1,t^i_1),\dots, (x_M,t^i_M)\} \subset \Omega \times [0,T]$, and
therefore minimize the empirical loss function
\begin{equation} \label{eq: discreteloss}
    \hat{\mathcal{L}}(\hat{\OpG}) = \frac{1}{NM} \sum_{\substack{1 \le i \le N \\ y \in S_i}} \bigl\|u^{(i)}(y) - \hat{\OpG}(u_0^{(i)}, f^{(i)})(y)\bigr\|_2^2.
\end{equation}

\begin{figure*}[!t]
\begin{adjustbox}{width=\textwidth}
    \input{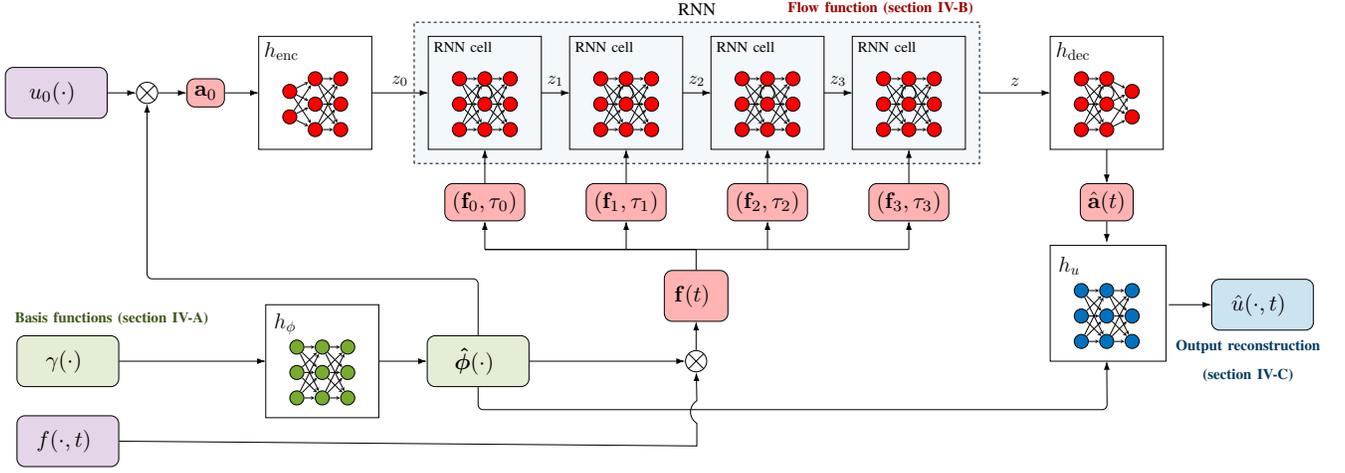}
\end{adjustbox}
%% After comments Kalle
\caption{RHYME-XT architecture. The model inputs ($u_0$ and $f$) are shown in {\textcolor{purple}{Purple}}. \textcolor{green}{Green} denotes the basis function generation, parameterized by the FNN $h_{\phi}$, where its inputs first pass through a random Fourier mapping $\gamma$. \textcolor{red}{Red} denotes the flow function architecture, taking in the projected inputs to generate the temporal coefficients.
\textcolor{blue}{Blue} denotes the output reconstruction, parameterized by the FNN $h_u$, which combines the basis functions and temporal coefficient to give the approximated solution.
Specifically, the inputs $u_0$ and $f$, are projected using the basis functions to obtain $\mathbf{a}_0$ and $\mathbf{f}(t)$, respectively. The projected initial condition is mapped to $z_0$ by the encoder $h_\mathrm{enc}$, after which it is propagated in time through the RNN with input parameters \begin{math} \{\mathbf{f}_k, \tau_k\}_{k=0}^{k_t}\end{math}, with $k_t=3$ in this example. The approximated temporal coefficients $\hat{\mathbf{a}}(t)$ are obtained by passing the output RNN state $z$ through the decoder $h_\mathrm{dec}$. These temporal coefficients are combined with the basis functions in the output reconstruction network to obtain the approximated output $\hat{u}(\cdot,t)$.}
    \label{fig:architecture}
    \vspace{-1.5em}
\end{figure*}
\section{Galerkin Projection of PIDE Dynamics} \label{sec: finite dimension}
In this section, we use a Galerkin method to project the PIDE dynamics
onto a finite-dimensional subspace spanned by a finite set of basis functions,
yielding a finite-dimensional system of controlled ODEs.
To simplify the notation, we denote restriction of a function to a given time instant by a superscript, e.g., $u^t := u(\cdot,t)$.

Assume that the solution $u^t$ of \eqref{eq:PIDEsfamily} belongs for each $t \geq 0$
to a real separable Hilbert space $\mathbb{H}$ with the standard inner product
$\langle g, h\rangle := \int_\Omega g(x)h(x) \mathrm{d}x$, ${g, h \in \mathbb{H}}$
and let ${\{\phi_k \in \mathbb{H} \mid k\in\mathbb{N}\}}$ be an orthonormal basis for $\mathbb{H}$.
The projection $\pi_r u^t$ of the solution $u^t$ onto
the subspace spanned by the basis functions $\{\phi_1, \dots, \phi_r\}$ is given by
\begin{equation} \label{eq:innerproduct3}
    \pi_r{u}^t = \sum_{n=1}^{r} a_n(t)\phi_n = \mathbf{a}(t)^\top \boldsymbol{\phi},
\end{equation}
where the coefficients ${a_n(t) = \langle \phi_n, u^t\rangle}$, 
${\mathbf{a} = \mathrm{col}(a_1, \dots, a_r)}$
and ${\boldsymbol{\phi} = \mathrm{col}(\phi_1, \dots, \phi_r)}$.

Applying Galerkin projection on \eqref{eq:PIDEsfamily} with \eqref{eq:innerproduct3},
we have
\begin{equation}
    \Bigl\langle \phi_n,\,
    {\dot{\mathbf{a}}(t)^\top\bphi}
    \Bigr\rangle= \Bigl\langle \phi_n, \OpF\!\left[
        \mathbf{a}(t)^\top \boldsymbol{\phi}
    \right] + f^t
    \Bigr\rangle,
\end{equation} for $n=1,2,...,r$. Additionally, $a_n(0) = \langle \phi_n, u_0 \rangle$.

Due to the orthonormality of the basis functions, this results in the following system of $r$ nonlinear ordinary differential equations (ODEs) 
\begin{equation} \label{eq:systemofodes3rrr}
    \dot{a}_n(t) = \left\langle \phi_n,
        \OpF\!\left[\mathbf{a}(t)^\top \bphi\right]
    \right\rangle + \langle \phi_n, f^t \rangle, \ n = 1, \dots, r.
\end{equation} 
The  state-space form 
\begin{equation} \label{eq:systemofodesv2}
\begin{aligned}
\dot{\mathbf{a}}(t) &= \mathbf{H}(\mathbf{a}(t)) + \mathbf{f}(t) \\
\mathbf{a}(0) &= \mathbf{a}_0
\end{aligned}
\end{equation}
may be derived from~\eqref{eq:systemofodes3rrr} by setting
\begin{equation} \label{eq: state space full}
\begin{aligned} 
    \mathbf{H} &= \mathrm{col}(H_1, \dots, H_r) \\
    \mathbf{f}(t) &= \mathrm{col}(\langle\phi_1,  f^t \rangle, \dots, \langle \phi_r,  f^t \rangle) \\
    \mathbf{a}_0 &= \mathrm{col}(\langle\phi_1,u_0\rangle, \dots, \langle \phi_r,u_0\rangle),
    % \mathbf{a}_0 &= \mathrm{col}(a_1(0), \dots, a_r(0)),
\end{aligned}
\end{equation}
with $H_n (\mathbf{a}(t))= \left\langle \phi_n,
        \OpF\!\left[\mathbf{a}(t)^\top \bphi\right]
    \right\rangle$.
Hence, if $\mathbf{a}(t)$ is a trajectory of~\eqref{eq:systemofodesv2}, then the projection of the corresponding solution $u$ of \eqref{eq:PIDEsfamily} may be computed as $\pi_r{u}(x, t) = \sum_{n=1}^r \phi_n(x)a_n(t)$.

\section{RHYME-XT Architecture} \label{sec: model approximation with Galerkin Projection}
In this section, we propose the RHYME-XT neural operator architecture to solve the learning problem introduced in section~\ref{sec: problem formulation}.
The architecture consists of three components, illustrated in Fig.~\ref{fig:architecture}. First, we use an FNN for approximating a set of basis functions. Secondly, we employ a flow function-based architecture to approximate the solution of the ODEs~\eqref{eq:systemofodesv2}.
Finally, these are combined using a second FNN to obtain the approximated solution at an arbitrary space-time point in $\Omega \times [0, T]$.

\subsection{Basis function architecture}
% We parameterize the basis functions by a neural network.
As introduced in \cite{TancikEtAl20}, to enhance the learning of high-frequency functions with low-dimensional input domains, we first pass the input $x$
through the random Fourier feature mapping 
\begin{equation}
\gamma(x) =
\begin{bmatrix}
\cos(\mathbf{B}x) \\
\sin(\mathbf{B}x)
\end{bmatrix},
\end{equation}
where $\mathbf{B} \in \mathbb{R}^{L\times d}$ is a matrix with entries sampled independently from a zero-mean Gaussian distribution.
The standard deviation of this distribution and the number of sampled features $L$ are hyperparameters.
Thus, letting $h_\phi: \Reals^L \to \Reals^r$ denote the FNN mapping the output of
$\gamma$ to the basis function values,
the overall representation of the basis functions can be given as
\begin{equation}\label{eq:basis-fnn}
    \boldsymbol{\hat{\phi}}(x) = h_\phi(\gamma(x)).
\end{equation}

\subsection{Flow function architecture}
We now consider the approximation of solutions of the system~\eqref{eq:systemofodesv2}.
We assume that the input functions $f$ in~\eqref{eq:PIDEsfamily} under consideration are measurable and uniformly bounded
\begin{equation}\label{eq:f-bounded}
    f \in \left\{
        g \in L_\infty(\Omega\times[0, T]) \mid \|g\|_\infty < M
    \right\}
\end{equation}
for some constant $M > 0$.
Furthermore, $f$ is assumed to be piecewise constant in the time variable with period $\Delta$, i.e.,
\begin{equation} \label{eq:f-piecewise-constant}
        f(\cdot,t) = f_k, \qquad k\Delta \leq t < (k+1)\Delta, \ k = 0, 1, \dots,
\end{equation}
where $f_k: \Omega \to \Reals$.
This implies that, in \eqref{eq:systemofodesv2}, the projected inputs $\mathbf{f}$ satisfy
\begin{math}
{\mathbf{f}(t) = \mathbf{f}_k \in \Reals^r}
\end{math}
for ${k\Delta \leq t < (k+1)\Delta}$.
Furthermore,~\eqref{eq:f-bounded} implies that 
\begin{equation*}
    \|\mathbf{f}_k\|^2_2 = \sum_{n=1}^r \langle f^{k\Delta},\phi_n\rangle^2 
        \leq \sum_{n=1}^r \|f^{k\Delta}\|_{\mathbb{H}}^2 \leq r M^2|\Omega|.
\end{equation*}
We thus define the set of projected inputs $\mathbb{F}$ to be the set of functions $\mathbf{f}: [0, T] \to \mathcal{F}$ satisfying~\eqref{eq:f-piecewise-constant},
where $\mathcal{F} \subset \Reals^r$ is the ball of radius $M\sqrt{r|\Omega|}$
centered at the origin.

Assuming that under the inputs in $\mathbb{F}$ the system~\eqref{eq:systemofodesv2} has an invariant set $\mathcal{A} \subset \Reals^r$, we may define the map
${\varphi: [0, T] \times \mathcal{A} \times \mathbb{F} \rightarrow \mathcal{A}}$,
such that $\varphi(t, \mathbf{a}_0, \mathbf{f})$ is the value of the state trajectory $\mathbf{a}(t)$
under the initial condition $\mathbf{a}_0 \in \mathcal{A}$ and input $\mathbf{f} \in \mathbb{F}$.
We call $\varphi$ the flow function of the system.

At this stage, we may directly apply the architecture proposed in~\cite{AguiarEtAl23}.
Fixing $(t, \mathbf{a}_0, \mathbf{f})$, we now describe how to compute the approximated
flow value $\hat\varphi(t, \mathbf{a}_0, \mathbf{f})$.
First, we introduce an encoder FNN denoted by~$h_{\mathrm{enc}}$ that maps the initial state to a feature space $\mathcal{Z}$. Denoting by $f_{\mathrm{RNN}}$ the mapping applied at each step by the RNN, we have the sequence of hidden states
\begin{equation} \label{eq:encoder}
\begin{aligned} 
    z_0 &= h_{\mathrm{enc}}(\mathbf{a}_0) \\
    z_{k+1} &= f_{\mathrm{RNN}}(z_k, \mathbf{f}_k, \tau_k), \ k = 0, \dots, k_t,
\end{aligned}
\end{equation}
where $z_i \in \mathcal{Z}$ are the hidden states of the RNN, $k_t = \lfloor t/\Delta\rfloor$ is the number of input values in $[0, t]$ and the $\tau_k$
are the fraction of time that the $k$th input value $\mathbf{f}_k$ is in effect,
that is $\tau_k = \min\{\frac{t - k\Delta}{\Delta}, 1\}$.
To enforce time continuity, the last two RNN states are combined as
\begin{equation}\label{eq:flow:interp}
    z = (1-\tau_{k_t})z_{k_t} + \tau_{k_t} z_{k_t+1}.
\end{equation}
Finally, $z$ is mapped to the state space $\mathcal{A}$ using a decoder FNN,
denoted by $h_{\mathrm{dec}}$, so that the approximated flow $\hat{\varphi}$ at time $t$ is given by
\begin{equation} \label{eq:decoder}
    \hat{\mathbf{a}}(t) = {\hat{\varphi}}(t,\mathbf{a}_0,\mathbf{f}) = h_{\mathrm{dec}}(z).
\end{equation}

\subsection{Output reconstruction}
Following \cite{SeidmanEtAl22}, rather than directly applying the linear basis expansion~\eqref{eq:innerproduct3}, we use a nonlinear mapping for improved performance.
Specifically, the predicted output, denoted by $\hat{u}(x,t)$, is given by the Hadamard (elementwise) product between the coefficients and basis functions
\begin{equation} \label{eq:outputreconstructioneqn}
   \hat{u}(x,t) = h_u(\hat{\mathbf{a}}(t) \odot \boldsymbol{\hat{\phi}}(x)),
    %= g(\hat{a}_1(t)\hat{\phi}_1(x), \dots, \hat{a}_r(t)\hat{\phi}_r(x)),
\end{equation}
where $h_u$ is an FNN. 

\section{Learning Methodology} \label{sec: Learning-based Approach}
To learn the RHYME-XT parameters from data, we propose a two-step procedure.
In the first step, the basis functions are learned. In the second step, these functions are used to project the system inputs to learn the flow function and output function parameterizations.

\subsection{Training data for the basis functions}
We use proper orthogonal decomposition (POD)~\cite[ch.~3]{HolmesEtAl12}
to obtain training data for the basis function network, yielding a set of orthogonal basis modes that capture the dominant dynamics of the system. 
We sample the data~\eqref{eq:data} at fixed spatial points
$x_1, \dots, x_{N_x}$ and at uniformly spaced times $t_1, \dots, t_K$,
resulting in the snapshots 
\begin{equation*}
    \mathbf{u}^{(i)}_k = \begin{bmatrix}
        u^{(i)}(x_1,t_k) & \cdots & u^{(i)}(x_{N_x},t_k)
    \end{bmatrix}^\top,
\end{equation*}
where $k = 1, \dots, K$.
We then collect the snapshots for each trajectory in a matrix
$\mathbf{A}_i = \begin{bmatrix}
\mathbf{u}^{(i)}_1 \;\mathbf{u}^{(i)}_2 \; \ldots \;\mathbf{u}^{(i)}_{K}
\end{bmatrix}$.
Subsequently, we perform singular value decomposition (SVD) of the block matrix
$\mathbf{A} = \begin{bmatrix} \mathbf{A}_1, \dots, \mathbf{A}_N \end{bmatrix} \in \Reals^{N_x \times M}$ with ${M = KN}$.
The SVD results in 
   ${\mathbf{A} = \mathbf{U} \mathbf{\Sigma}\mathbf{V}^\top}$,
where ${\mathbf{U} \in \Reals^{N_x \times N_x}}$ and ${\mathbf{V} \in \Reals^{ M \times M}}$ are unitary matrices, and ${\boldsymbol{\Sigma} \in \mathbb{R}^{N_x \times M}}$ is a diagonal matrix of singular values.

We truncate this decomposition by retaining the first $r$ singular values and the corresponding left and right singular vectors, leading to a low-rank approximation 
${\mathbf{A} \approx \tilde{\mathbf{U}} \tilde{\mathbf{\Sigma}}\tilde{\mathbf{V}}^\top}$ with \begin{math}
         \tilde{\mathbf{U}} \in \Reals^{N_x \times r}
\end{math}, $\tilde{\mathbf{V}}^{\top} = \begin{bmatrix}
    \tilde{\mathbf{V}}^\top_1 & \dots & \tilde{\mathbf{V}}^\top_N
\end{bmatrix}$ and ${\tilde{\mathbf{V}}_i \in \Reals^{K \times r}}$.
In other words, for the snapshot at time $t_k$ we have
$\mathbf{u}^{(i)}_k \approx
        \tilde{\mathbf{U}}\tilde{\mathbf{\Sigma}}\tilde{\mathbf{V}}_i^\top e_k
    = \textstyle \sum_{n = 1}^r \sigma_n v^{(i)}_{kn} u_n$,
where $e_k$ is the $k$th basis vector, $\sigma_n$ and $v^{(i)}_{nk}$ are the entries of $\tilde{\mathbf{\Sigma}}$ and
$\tilde{\mathbf{V}}_i$, respectively, and $u_n$ are the columns of $\tilde{\mathbf{U}}$.
Comparing with~\eqref{eq:innerproduct3}, we see that the latter provide the optimal values of the basis functions at the selected spatial locations.

We train $\boldsymbol{\hat\phi}$ in~\eqref{eq:basis-fnn} by minimizing the loss function
\begin{equation}
\label{eq:loss}
    \mathcal{L}_{\phi}(\boldsymbol{\hat{\Phi}}) = ||\boldsymbol{\hat{\Phi}}-\tilde{\mathbf{U}}||_1 + ||\boldsymbol{\hat{\Phi}}^{\top} \boldsymbol{\hat{\Phi}}-I_r||_1,
\end{equation}
where the matrix $\boldsymbol{\hat{\Phi}} = \begin{bmatrix}
    {\boldsymbol{\hat\phi}}(x_1) & \dots & {\boldsymbol{\hat\phi}(x_{N_x})}
\end{bmatrix}^\top \in \Reals^{N_x \times r}$, i.e., the values of the basis function FNN at the aforementioned spatial locations.
The second term in \eqref{eq:loss} is used as a regularizer to promote orthonormality in the resulting basis functions.

\subsection{Flow function and output reconstruction}
To determine a surrogate for \eqref{eq:systemofodesv2}, we first use the basis functions \eqref{eq:basis-fnn} to compute $\mathbf{a}_0$ and $\mathbf{f}$ using the respective inner products in \eqref{eq: state space full}. These inner products can be numerically approximated by discretizing the inputs, and subsequently summing them in the spatial dimension with appropriate quadrature weights. Given the flow function output \eqref{eq:decoder} and the basis functions, \eqref{eq:outputreconstructioneqn} yields an approximate solution to \eqref{eq:PIDEsfamily} at arbitrary evaluation points.
For each trajectory in the data \eqref{eq:data}, we define a set of evaluation points $S_i$, and sample the data at the points in this set to obtain the corresponding ground-truth outputs.
The flow function and output reconstruction networks are then trained simultaneously by minimizing the empirical loss \eqref{eq: discreteloss}.

\section{Numerical Results}  \label{sec: Results}
In this section, we demonstrate the performance of RHYME-XT on a network of neurons.

\subsection{Data generation}
We consider the data model~\eqref{eq:neural_field} defined on a one-dimensional finite domain $\Omega = [-10,10]$ and $\theta=1$. The weighting function is given by a difference of Gaussians,
\begin{equation} \label{eq: mexhat}
    k(x) = 3 \exp\left({\tfrac{1}{2}\left(\tfrac{x}{1.5}\right)^2}\right) - 1.5 \exp\left({\tfrac{1}{2}\left(\tfrac{x}{3}\right)^2}\right) - 0.2,
\end{equation} and the firing rate function is a sigmoid,
\begin{equation*}
    h(x) = \left({1+\exp \left(-1000 \left(x-\theta\right)\right)}\right)^{-1}.
\end{equation*}

We generate $N=1000$ trajectories as in~\eqref{eq:data} on $[0,T]=[0,50]$ with temporal and spatial discretizations of $\Delta t = 0.025$ and $\Delta x=0.025$, respectively, using a forward Euler scheme for time integration. The spatial convolutions were computed using a fast Fourier transform and its inverse, assuming periodic boundary conditions.
For each trajectory, we sample the initial conditions $u_0^{(i)}$
i.i.d. according to
\begin{equation*}
    u_0(x) = \textstyle\sum_{i=1}^{2} A_i {\exp}{\left(-\tfrac{\left(x-\mu_i\right)^2}{2\sigma_i^2}\right)},
\end{equation*}
with uniformly distributed parameters
$A_i \sim \mathrm{Unif}([1, 5])$ and $\sigma_i \sim \mathrm{Unif}([0.5, 2.5])$.
The input functions $f^{(i)}$ have period $\Delta=0.5$ and are sampled i.i.d. according to 
\begin{equation*} 
\begin{aligned}
    f_{10k} (x) &= A\exp{\left(-\tfrac{\left(x-\mu\right)^2}{2\sigma^2}\right)}, \quad k \geq 0 
\end{aligned}
\end{equation*}
and $f_{10k+j}(x) = f_{10k}(x)$, $j = 1, \dots, 9$, i.e., the input value changes every 5~time units.
The parameters $A$ and $\sigma$ are distributed as above and $\mu \sim  \mathrm{Unif}([-10, 10])$.

\subsection{Model settings}
\textbf{RHYME-XT}. For the basis functions, $r=50$ was used. The FNN used to model these consists of four layers with 100 neurons each.
The number of random Fourier features is $L=128$, with standard deviation $0.5$.
In the flow function architecture, the RNN is a long short-term memory network with 250 hidden states. The encoder FNN consists of two hidden layers with 500 neurons each, while the decoder FNN consists of three hidden layers with 250 neurons each. The FNN used for the output reconstruction has two layers with 50 neurons. All networks use $\tanh$ activations. 

\textbf{Baseline}. The DeepONet model is configured to have the same number of trainable parameters as our proposed method. As the inputs $f_k$~\eqref{eq:f-piecewise-constant} change every 5 units of time, we consider the initial condition and that (constant in time) input to be the inputs to the branch network.
The trunk network, encoding the coordinates of the output, takes as input the location $x$ and time $t$ of the output. The output is then given as the same nonlinear output reconstruction \eqref{eq:outputreconstructioneqn} as used in our proposed method, with $r=700$ modes.
Following~\cite{WangPerdikaris23}, when the input $f$ changes in time, the output of the model at the previous timestep is used as the new initial condition that is passed to the branch network.

\subsection{Results}
The data was split into a train, validation, and test set with a 70\%-20\%-10\% split. In the first step of training, the full set of temporal points in the trajectories was used. In the second step, a subset of~200 time points was selected using Latin hypercube sampling.
In both steps, we use $N_x=100$ uniformly spaced spatial points.
We use the stochastic gradient descent algorithm \texttt{Adam}.
For the second step, we use a batch size of $128$ and the initial learning rate was set to $1.2 \times 10^{-4}$, which was reduced by a factor of~2 whenever the validation loss did not decrease for 5~consecutive epochs.
The training was stopped when the validation loss did not decrease for 30~consecutive epochs.

\begin{figure}[!t]
    \centering
    \includegraphics[width=\columnwidth]{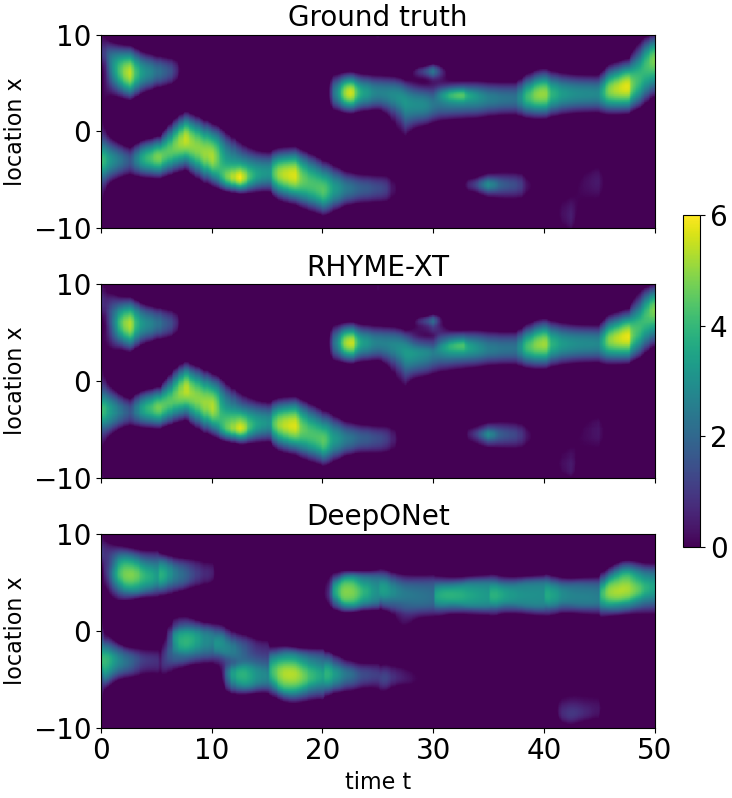}
    \caption{Visualization of a trajectory of our proposed method RYHME-XT versus the baseline DeepONet. The relative $\ell^2$ errors of the trajectory are $0.0571$ for RHYME-XT and $0.4810$ for DeepONet, respectively.}
    \label{fig:results_don}
    %\vspace{-1em}
\end{figure}
\begin{figure}[!t]
    \centering
    \includegraphics[width=\columnwidth]{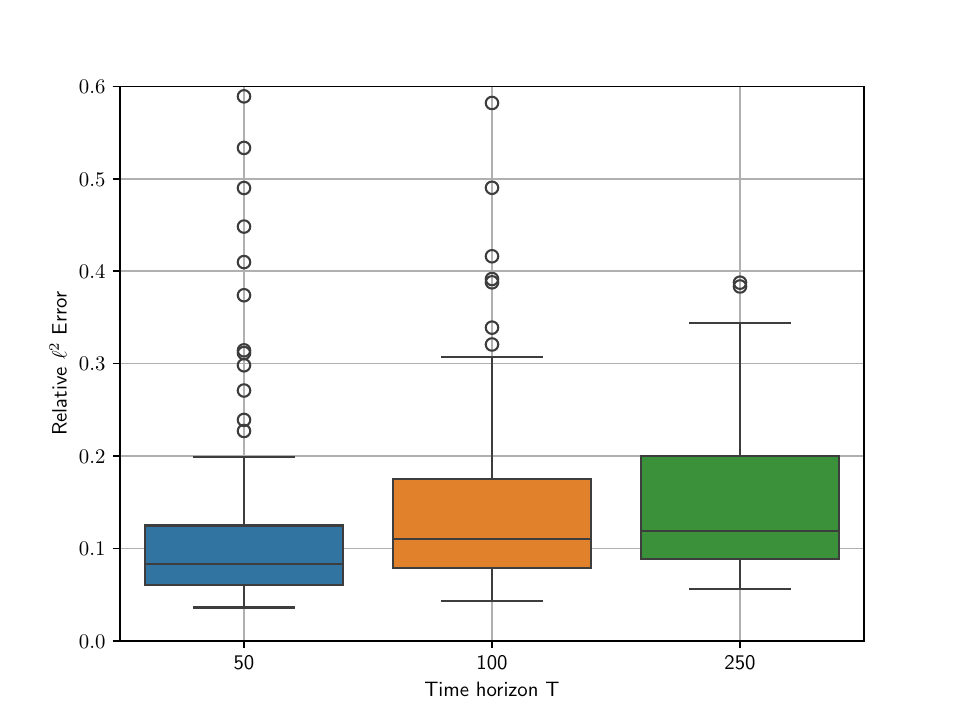}
    \caption{Box plot of the relative $\ell^2$ error for different time horizons $T$, with the model being trained on trajectories with $T=50$.}
    \label{fig:boxplotT}
    % \vspace{0.3em}
    % \hrule
    % \vspace{-2em}
    %\vspace{-1.5em}
\end{figure}

\textbf{Comparison with baseline}. Over the test trajectories, our method achieves a relative $\ell^2$ error of $0.1284 \pm 0.1287$,  while the baseline (DeepONet) achieves a relative $\ell^2$ error of  $0.7115 \pm 0.1078$. This difference in performance can be seen in Fig.~\ref{fig:results_don}, where it is noticeable that our model is able to capture the fine-grained details in the trajectory.% in comparison with the baseline.

\begin{figure*}[!t]
    \centering
    \includegraphics[width=5.9in]{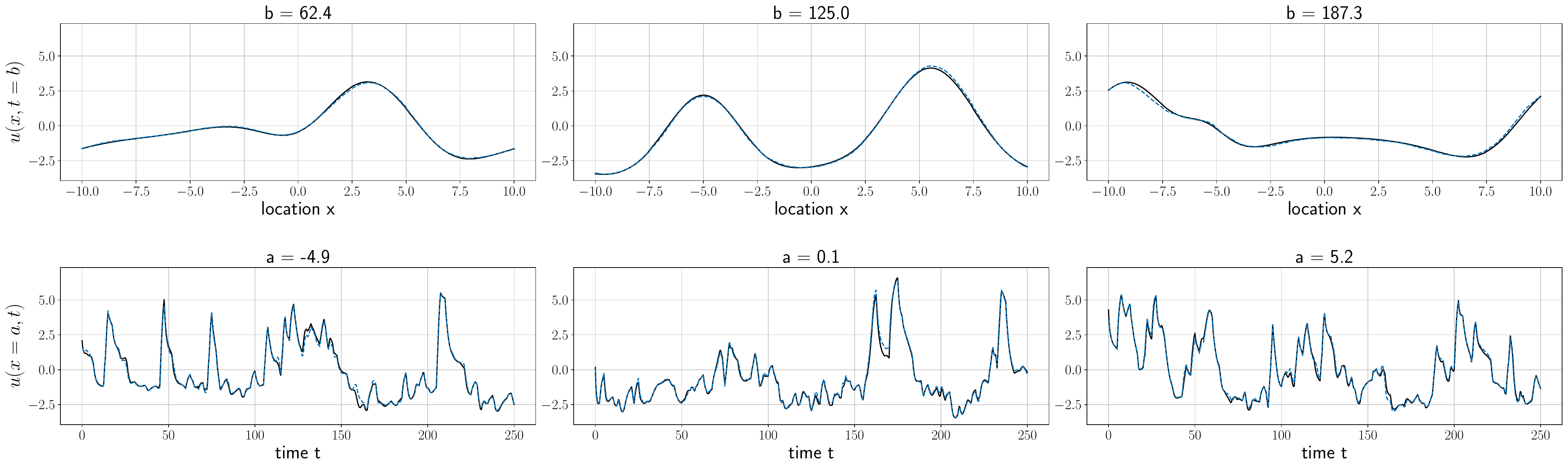}
    \caption{Visualization of a trajectory generated with $T=250$ versus model predictions of our proposed method RHYME-XT, trained using trajectories with $T=50$. Ground truth: (\texttt{-}), RHYME-XT:  (\textcolor{blue}{\texttt{--}}). The relative $\ell^2$ error of this trajectory is $0.1173$.}
    \label{fig:T200}
    \vspace{-1.5em}
\end{figure*}

\begin{figure}[!t]
    \centering
    \includegraphics[width=2.8in]{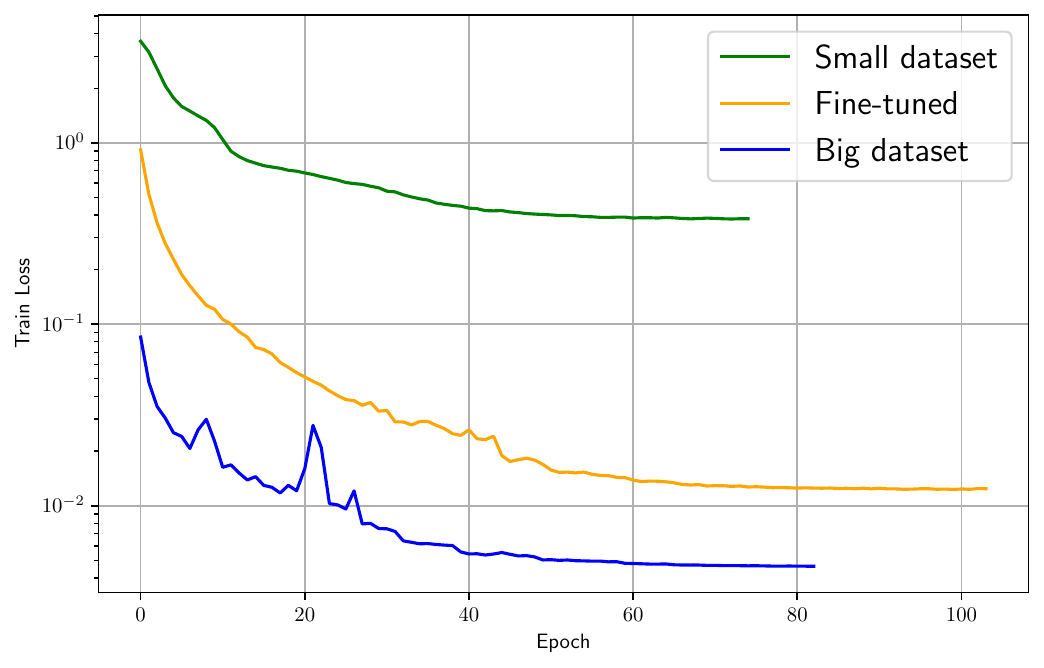}
    \vspace{-.7em}
    \caption{Training loss trajectory for different training scenarios.}
    \label{fig:boxplotTransferlearning}
    \vspace{-1.5em}
\end{figure}

\textbf{Prediction over large time horizons}.
We investigate the performance of our model, trained on a temporal horizon of $T=50$, over longer time horizons.
Specifically, we consider 100 trajectories with temporal horizons $T=100$ and $T=250$, respectively.
Fig.~\ref{fig:boxplotT} shows a box plot of the corresponding losses for each value of $T$.
We observe that the loss increases slightly from $T=50$ to $T=100$, with the mean $\ell^2$ error going from $0.1283$ to $0.1415$, but shows no significant increase beyond that step, with a mean error of $0.1546$ for a time horizon $T=250$, indicating that there is no significant error propagation.
This can also be observed in Fig.~\ref{fig:T200}, where some slices in time and space for a trajectory with $T=250$ are plotted.
%%% PDF
\textbf{Transfer learning}.
Finally, we consider fine-tuning an already trained model on a small amount of data from a different system, to obtain a surrogate of the new system.
To evaluate this, we consider three training datasets: (i) a high amount of data generated as before, (ii) a small amount of data, i.e., $2\% $ of the amount used in (i), generated as before, and (iii) a high amount of data, but generated instead with a Gaussian weighting function~$k$.
Specifically, we first train the model on dataset~(iii).
The learned parameters of the flow function and the output reconstruction networks are then used to initialize the model that is subsequently trained on the small dataset~(ii).
The results are shown in Fig.~\ref{fig:boxplotTransferlearning}, where `Fine-tuned' is the curve for the resulting model, `Small dataset' corresponds to training only on dataset~(ii), while `Big dataset' is a model trained on dataset~(i).
The results indicate that it is indeed feasible to adapt a model to a new system
by fine-tuning prior models (more details can be found in the code repository%
\footnote{\url{https://github.com/Marijn-dev/RHYME-XT}}).

\section{Conclusion and Perspectives} \label{sec: Conclusion}
We presented an operator learning framework to learn the surrogate model of excitable spatiotemporal control systems from trajectory data.
In a Galerkin projection-based framework, we used a learning approach to leverage the approximation capabilities of neural networks.
Our method shows improved results compared to the baseline on a neural field model. Moreover, our method retains model accuracy with respect to extrapolation in time. Furthermore, we demonstrated the ability to transfer dynamics knowledge by fine-tuning on small amounts of data. %, with no significant increase in the error as the time horizon is increased.

There are many potential avenues of research to improve upon and develop the method. For example, employing a Petrov-Galerkin projection, that is, using different basis functions in the projection of the solution and the dynamics, may provide benefits.
Regarding experimental evaluation, a study of the performance for nonlinear PDE systems and a more comprehensive comparison with other state-of-the-art methods, considering model accuracy and computational trade-offs, should be performed.
Finally, theoretical approximation properties of the proposed architecture should be investigated.

\newcommand{\DecoderBlock}[1]{%
  \node[
    draw,
    rounded corners,
    minimum width=2.2cm,
    minimum height=1.6cm,
    align=center
  ] (#1) {
    Basis Functions FNN\\[2pt]
    \begin{tikzpicture}[scale=0.4]
      % \foreach \x in {0,1,2} {
      %   \foreach \y in {0,1,2} {
      %     \fill[green!60] (\x,\y) circle (2pt);
      %   }
      % }
    \end{tikzpicture}
  };
}
\vspace{-0.5em}
\section*{Acknowledgments}
This research is supported by Swedish Research Council Distinguished Professor Grant 2017-01078
Knut and Alice Wallenberg Foundation Wallenberg Scholar Grant.
The computations were enabled by resources provided by the National Academic Infrastructure for Supercomputing in Sweden (NAISS), partially funded by the Swedish Research Council through grant agreement no. 2022-06725.

% \end{thebibliography}
\bibliographystyle{IEEEtran}
\bibliography{references}
\end{document}